# Margin Distribution Controlled Boosting

Guangxu Guo and Songcan Chen[*]

College of Computer Science & Technology, Nanjing University of Aeronautics & Astronautics, Nanjing, China 210016.

http://parnec.nuaa.edu.cn

**Abstract**— Schapire's margin theory provides a theoretical explanation to the success of boosting-type methods and manifests that a *good* margin distribution (MD) of training samples is essential for generalization. However the statement that a MD is good is vague, consequently, many recently developed algorithms try to generate a MD in their goodness senses for boosting generalization. Unlike their *indirect* control over MD, in this paper, we propose an alternative boosting algorithm termed Margin distribution Controlled Boosting (MCBoost) which *directly* controls the MD by introducing and optimizing a key adjustable margin parameter. MCBoost's optimization implementation adopts the column generation technique to ensure fast convergence and small number of weak classifiers involved in the final MCBooster. We empirically demonstrate: 1) AdaBoost is actually also a MD controlled algorithm and its iteration number acts as a parameter controlling the distribution and 2) the generalization performance of MCBoost evaluated on UCI benchmark datasets is validated better than those of AdaBoost, L2Boost, LPBoost, AdaBoost-CG and MDBoost.

## 1  Introduction

Boosting is a technique to create a strong classifier by linearly combining a set of weak classifiers. A weak classifier is one which performs just slightly better than random guessing. In contrast, a strong classifier has better accuracy. In the learning process for most of two-class boosting algorithms, there is a set of weights on training examples maintained and sent to the oracle in each iteration to obtain a base classifier, the algorithms then update the weights so that larger weights are put on those misclassified examples.

AdaBoost [1] is the most popular boosting algorithm in the machine learning community and has achieved

[*] Corresponding author: Tel: +86-25-84892956 and E-mail: s.chen@nuaa.edu.cn (S.C. Chen)

great success due to its excellent generalization and ease of implementation. Friedman et.al. [2] interpreted the working mechanism of AdaBoost, from the statistical point of view, as a stagewise optimization for fitting an additive logistic regression model. Following this interpretation, there were many algorithms and theoretical analyses (for example [3], [4], [5]) proposed successively. Despite their success, many unanswered questions still remain [6]. An interesting empirical observation is that AdaBoost seems to be immune to overfitting in many datasets [7], [8]. The generalization error keeps decreasing as its iteration number increases, even after the training error reaches zero. Many researchers have been attracted by this phenomenon and try to give an interpretation.

Schapire et.al. [9] explained this phenomenon through large margin theory and derived margin-based upper bounds on the generalization error of boosting algorithms, from which they concluded that the larger the margins, the lower the generalization error. Besides, they also empirically demonstrated that AdaBoost converges to a MD in which most examples have large margins. Soon after, Breiman [10] presented tighter margin bounds which depend on the minimum margin and designed a variant of AdaBoost called arc-gv. In his experiments, arc-gv could always achieves larger minimum margins than AdaBoost, but performs worse according to generalization. While Breiman also found that the MDs got by arc-gv are always better than that by AdaBoost which totally violates the margin theory. Thus, he concluded that neither the MD nor the minimum margin dominates generalization. Later, Reyzin and Schapire [11] took a closer look at Breiman's experiments and found that the base classifiers used in arc-gv are more complex. Then they re-ran the experiments by fixing the complexity of the base classifiers and observed that arc-gv still produces worse performance and larger minimum margins, but actually poor MDs. As a result, the arc-gv's inferiority is re-explained by the margin theory.

Motivated by support vector machines (SVMs), LPBoost [12, 13] was designed to maximize the minimum margin. For linearly separable data, a hard-margin LPBoost works well, while for linearly inseparable data, a soft-margin LPBoost, which can be explained by maximizing the average of the k smallest margins [14], performs relatively more robustly. Despite its simplicity and nature of motivation, LPBoost actually has poor generalizations in most data sets, and has been observed to over-weight those examples with the minimum margin. Recently, there were many algorithms, such as softBoost [15] and ERLPBoost [16], designed to overcome this problem.

Most recently, the average and the variance of margins begun attracting a concern again typically due to [11], in which Reyzin and Schapire found that arc-gv maximizes the minimum margin by sacrificing margins of overmany examples, consequently, they conjectured that maximizing the average or median margin may be fruitful. Shen and Li [17] proved that AdaBoost approximately maximizes the average margin rather than the minimum margin under some assumptions. They designed an algorithm termed MDBoost [18] which directly maximizes the average margin and minimizes the margin variance simultaneously, and use cross-validation (CV) to balance the weights of the average margin and the margin variance. In addition, motivated by Bernstein's inequality, Shivaswamy and Jebara [19] designed an algorithm called EBBoost which improves the performance of AdaBoost by minimizing a convex function added with the empirical variance of the exponential loss. Both MDBoost and EBBoost take the margin variance into account and outperform AdaBoost.

It has been profoundly believed that MD is closely related to generalization. However, how to measure its goodness is still an unanswered question. In this paper we proposed a totally corrective boosting algorithm termed MCBoost by *directly* controlling the MD of training samples. Its motivation is simple and intuitive as well. Specifically, we control the MD by minimizing the sum of the squared differences between the margins and an introduced adjustable parameter which plays a (desirable) margin role. In this way, the MD changes with the parameter and finally the best one is determined by CV.

Different from the above-mentioned algorithms producing MDs by optimizing their own measures (for example, the minimum or average margins) which characterize the goodness of a MD to different degree, our MCBoost *directly* produces MDs by adjusting the margin parameter and only takes the commonly-used validation error as the goodness measure. MCBoost can be established by solving a quadratic program and its optimization is implemented with the column generation (CG) technique. Finally our contributions are summarized as follows:

1) We propose a novel MCBoost algorithm which directly controls the MD, and use the CV to achieve best performance.

2) We empirically demonstrate that AdaBoost can also be regarded as a MD-controlled algorithm and its iteration number acts as a parameter to control the MD.

3) The generalization performance of MCBoost evaluated on UCI benchmark datasets is validated better than those of AdaBoost, L2Boost, LPBoost, AdaBoost-CG and MDBoost.

The rest of this paper is organized as follows: Section 2 introduces some notations and basic definitions. In Section 3 we present the motivation and the main idea of MCBoost. In Section 4, the dual of MCBoost is derived, through which we design a totally corrective boosting algorithm using CG. The result of experiments on UCI benchmark datasets is shown in Section 5 and we conclude the paper in Section 6.

## 2 Notations and Basic Definitions

Throughout this paper, we denote a column vector with a bold face low-case letter (e.g. **u** and **w**) and a matrix with an upper-case letter (e.g. $H$), use $\mathbf{u}^T$, $\mathbf{w}^T$, $H^T$ to denote their corresponding transposes. $\mathbf{u}^T\mathbf{w}$ denotes the inner product. The $i$th row (the $j$th column) of $H$ is $H_{i:}$ ($H_{:j}$). We use **1** (or **0**) to denote the column vector with all elements being 1 (or 0).

We focus on two-class problem and are given a training dataset with $M$ labeled examples $(\mathbf{x}_i, y_i)$, $i = 1...M$, where the predictors $\mathbf{x}_i$ are in domain X and $y_i \in \{-1, +1\}$. Let $h(\cdot): X \to [-1, +1]$ be a base classifier from a set $H = \{h_1,...,h_N\}$, and if $h_j \in H$ then $-h_j \in H$. Let $H$ be a matrix of size $M \times N$ and its entry $H_{ij} = h_j(\mathbf{x}_i)$. Thus $H_{i:}$ represents the classification results of all the base classifiers on $\mathbf{x}_i$, and $H_{:j}$ encodes the base classifier $h_j$.

Boosting builds a strong classifier by linearly combining a set of weak classifiers in terms of

$$F(\mathbf{x}) = \sum_{j=1}^{N} w_j h_j(\mathbf{x}), w_j \geq 0, \qquad (1)$$

where $w_j$ is the weight of $h_j$ and let $\mathbf{w} = [w_1,...,w_N]^T$. For $(\mathbf{x}_i, y_i)$ its normalized margin is defined as

$$\rho_i = \frac{y_i \sum_{j=1}^{N} w_j h_j(\mathbf{x}_i)}{\sum_{j=1}^{N} w_j} = \frac{y_i H_{i:} \mathbf{w}}{\|\mathbf{w}\|_1}. \qquad (2)$$

Loosely speaking, a margin can be viewed as a measure of how much this example is correctly classified. A training example predicted perfectly by all the base classifiers has the margin of 1 and that always predicted incorrectly has the margin of -1. $\mathbf{x}_i$ is correctly predicted *iff* $\rho_i \succ 0$. Ideally, all training examples should be correctly classified i.e. $\rho_i \succ 0, \forall i$. However, this is impossible in most real-world datasets duo to the existence of the noisy data (for example, overlapping class distribution, outliers and mislabeled patterns). Schapire's margin theory [9] indicates that maximizing the overall margin can improve the classification accuracy of a classifier.

## 3 Motivation and Main Idea

Our motivation mainly comes from both L2Boost [3] and AdaBoost-CG [17]. AdaBoost-CG modifies AdaBoost by constraining $\mathbf{1}^T\mathbf{w}$ and is empirically demonstrated better than AdaBoost [18]. While we modify

L2Boost by a similar constraint, nevertheless our algorithm derived is more intuitive.

A two-class L2Boost [3] works by minimizing the least squares loss function, which is equivalent to

$$\min_{\mathbf{w}} \sum_{i=1}^{M} (1 - y_i H_{i:} \mathbf{w})^2, \ s.t. \ \mathbf{w} \geq 0 \cdot \quad (4)$$

Define $\tilde{\rho}_i = y_i H_{i:} \mathbf{w}$ as the *unnormalized* margin and $\rho_i = (y_i H_{i:} \mathbf{w})/\|\mathbf{w}\|_1$ as the *normalized* margin with respect to (w.r.t.) $\mathbf{x}_i$. Throughout this paper, we use *margin* to denote *normalized* margin unless otherwise specified. The ideal minimum of (4) is 0 which happens iff $\tilde{\rho}_i = 1, \forall i$, i.e., L2Boost controls the *unnormalized* margins to a desired value of 1. However what we really concern is the (*normalized*) margins since the *unnormalized* margin can be affected by the scaling of $\mathbf{w}$. Next, in order to make clear of the MD produced by L2Boost, we reformulate (4) into (5) using the normalization of $\mathbf{w}$:

$$\min_{\mathbf{w}} \left[ \sum_{i=1}^{M} (\frac{1}{\|\mathbf{w}\|_1} - \frac{y_i H_{i:} \mathbf{w}}{\|\mathbf{w}\|_1})^2 \right] \|\mathbf{w}\|_1^2, \ s.t. \ \mathbf{w} \geq 0 \cdot \quad (5)$$

In [3], L2Boost has been shown to be numerically convergent, meaning that $1/\|\mathbf{w}\|_1$ will eventually stabilize to some value, hence the first sum-term of (5) can be roughly interpreted as making a MD approach to the value. However the MD achieved by L2Boost is not necessarily good, as a result, which motivates us to conceive that if the value of $1/\|\mathbf{w}\|_1$ is controllable or adjustable, the generalization performance is more likely improved.

Employing the trick in [17], for every fixed value of $\|\mathbf{w}\|_1 = 1/E$, problem (5) becomes the simplified one of (6):

$$\min_{\mathbf{w}} \sum_{i=1}^{M} (E - y_i H_{i:} E \mathbf{w})^2, \ s.t. \ \mathbf{w} \geq 0 \ and \ \|\mathbf{w}\|_1 = \frac{1}{E} \cdot \quad (6)$$

We substitute $E\mathbf{w}$ in (6) for $\mathbf{w}$, then our optimization problem can be further simplified to:

$$\min_{\mathbf{w}} \sum_{i=1}^{M} (E - y_i H_{i:} \mathbf{w})^2, \ s.t. \ \mathbf{w} \geq 0 \ and \ \|\mathbf{w}\|_1 = 1 \cdot \quad (7)$$

Equivalently,

$$\min_{\mathbf{w}} \sum_{i=1}^{M} (\rho_i - E)^2, \ s.t. \ \rho_i = y_i H_{i:} \mathbf{w}, \mathbf{w} \geq 0 \ and \ \|\mathbf{w}\|_1 = 1. \quad (8)$$

Now the optimization problem (4) of L2Boost becomes the one of our MCBoost (i.e. (8)) imposed the constraint $\|\mathbf{w}\|_1 = 1/E$. One can obtain different MDs of MCBoost by changing the value of $E$ and then optimizing (8), thus $E$ can be viewed as a hyperparameter controlling the MD, or more specifically a desired margin value, for which we

can determine its optimum by CV.

For further analyzing (8), we decompose it into

$$\sum_{i=1}^{M}(\rho_i\text{-}E)^2 = \sum_{i=1}^{M}(\rho_i\text{-}\overline{\rho})^2 + M(\overline{\rho}\text{-}E)^2 + 2\sum_{i=1}^{M}(\rho_i\text{-}\overline{\rho})(\overline{\rho}\text{-}E) , \qquad (9)$$

where $\overline{\rho} = (\sum_{i=1}^{M}\rho_i)/M$ is the average margin. The last term of (9) is equal to zero, so:

$$\frac{1}{M}\sum_{i=1}^{M}(\rho_i - E)^2 = \frac{1}{M}\sum_{i=1}^{M}(\rho_i - \overline{\rho})^2 + (\overline{\rho} - E)^2 \qquad (10)$$

$$= \underbrace{\frac{1}{M}\sum_{i=1}^{M}(\rho_i - \overline{\rho})^2 - 2E\overline{\rho} + \overline{\rho}^2}_{\text{the objective function of MDBoost}} + E^2 . \qquad (11)$$

We find that the first term of (10) is exactly the variance of the margins and the second term is exactly the squared bias between the average margin and *E*. Therefore the MD achieved by MCBoost should have a small variance and simultaneously an average margin close to *E*. As can be seen from (11) that the objective function of MCBoost is *totally* different from that of MDBoost [18], MDBoost controls the MD *indirectly* by trading-off the variance and the minus average margin so as to boost generalization ability. In contrast, our MCBoost *directly* controls the average margin to a desired value, and meanwhile forces the margins to aggregate closely around the average margin by minimizing the variance term of (10).

In order to facilitate solving, we define a column vector $\mathbf{\rho} = [\rho_1,...,\rho_M]^T$ and return the original objective (8) and rewrite it as a compact version:

$$\min \mathbf{\rho}^T\mathbf{\rho} - 2E\mathbf{1}^T\mathbf{\rho},\ s.t.\ \rho_i = y_i H_{i:}\mathbf{w},\ \mathbf{1}^T\mathbf{w} = 1, \mathbf{w} \geq 0 . \qquad (12)$$

Obviously, problem (12) is a strictly convex quadratic programming (QP) w.r.t. **w** and thus has a unique global optimum, while the objective in [18] is not strictly convex and thus its optimal solution is not generally unique.

If $\mathcal{H}$ is finite and its size is supposed not very large, then (12) can be solved directly using a QP solver. However, in most cases, the number of possible base classifiers is too large to be directly treated by standard QP solvers. Thus as in [18], CG is instead used to solve such a large scale optimization problem, refer to [13, 18] for more details. In the next section, the dual of MCBoost is derived and based on which an algorithm is designed.

**4 The Dual of MCBoost**

In this section we compute the Lagrangian dual of (12) to develop our MCBoost. The Lagrangian of (12) is

defined as

$$L(\underbrace{\mathbf{w},\boldsymbol{\rho}}_{\text{primal}},\underbrace{\mathbf{u},\mathbf{q},r}_{\text{dual}}) = \boldsymbol{\rho}^{\mathrm{T}}\boldsymbol{\rho} - 2E\mathbf{1}^{\mathrm{T}}\boldsymbol{\rho} + \sum_{i=1}^{M} u_i(\rho_i - y_i H_{i:}\mathbf{w})$$

$$-\mathbf{q}^{\mathrm{T}}\mathbf{w} + r(\mathbf{1}^{\mathrm{T}}\mathbf{w} - 1), \mathbf{q} \geq \mathbf{0}. \tag{13}$$

Its dual can be derived by minimizing the Lagrangian w.r.t. the primal variables:

$$\inf_{\boldsymbol{\rho},\mathbf{w}} L = \inf_{\boldsymbol{\rho}} (\boldsymbol{\rho}^{\mathrm{T}}\boldsymbol{\rho} - 2E\mathbf{1}^{\mathrm{T}}\boldsymbol{\rho} + \mathbf{u}^{T}\boldsymbol{\rho}) +$$

$$\inf_{\mathbf{w}} (r\mathbf{1}^{\mathrm{T}} - \sum_{i=1}^{M} u_i y_i H_{i:} - \mathbf{q}^{\mathrm{T}})\mathbf{w} - r, \mathbf{q} \geq 0. \tag{14}$$

The Lagrangian is linear in **w**. Therefore, according to the KKT conditions, the term multiplied by the **w** must be zero, i.e.,

$$r\mathbf{1}^{\mathrm{T}} - \sum_{i=1}^{M} u_i y_i H_{i:} - \mathbf{q}^{\mathrm{T}} = 0. \tag{15}$$

Since $\mathbf{q} \geq 0$, we have

$$\sum_{i=1}^{M} u_i y_i H_{i:} \leq r\mathbf{1}^{\mathrm{T}}. \tag{16}$$

By differentiating $L$ w.r.t. $\boldsymbol{\rho}$ and zeroing the derivative, we get

$$\boldsymbol{\rho} = -\frac{\mathbf{u} - 2E\mathbf{1}}{2}. \tag{17}$$

By plugging (15) and (17) into (13), the dual problem can be derived as follows

$$\max -r - \frac{1}{4}(\mathbf{u} - 2E\mathbf{1})^{\mathrm{T}}(\mathbf{u} - 2E\mathbf{1})$$

$$s.t. \sum_{i=1}^{M} u_i y_i H_{i:} \leq r\mathbf{1}^{\mathrm{T}}, \tag{18}$$

equivalently, having

$$\min r + \frac{1}{4}(\mathbf{u} - 2E\mathbf{1})^{\mathrm{T}}(\mathbf{u} - 2E\mathbf{1})$$

$$s.t. \sum_{i=1}^{M} u_i y_i H_{i:} \leq r\mathbf{1}^{\mathrm{T}}. \tag{19}$$

Due to both the convexity of the objective function in (12) and the linearity of the constraints, the problem satisfies Slater's condition and thus the strong duality holds, leading to zero dual gap. Thus optimizing (12) is equivalent to optimizing (19) which can be solved using CG [13]. MCBoost is described in Algorithm 1.

**Algorithm 1.** MCBoost

**Input:** Training examples $(\mathbf{x}_i, y_i)$, $i = 1,...,M$; Termination threshold $\varepsilon \succ 0$; Hyperparameter $E$ and The maximum number of iterations $N_{\max}$.

**Initialization:** $u_i = 1/M$, $i = 1,...,M$.

**Do for** $t = 1 : N_{\max}$

**(a)** Send $\mathbf{u}$ to oracle and obtain a new base classifier $h_t$.

**(b)** If $t > 1$ and $\sum_{i=1}^{M} u_i y_i h_t(\mathbf{x}_i) < r + \varepsilon$ then break.

**(c)** Update $\mathbf{u}$ and $r$, $[\mathbf{u}, r] = \arg\min_{\mathbf{u},r} r + \frac{1}{4}(\mathbf{u} - 2E\mathbf{1})^{\mathrm{T}}(\mathbf{u} - 2E\mathbf{1})$

$$s.t. \sum_{i=1}^{M} u_i y_i h_j(\mathbf{x}_i) \leq r, \, j = 1,...,t.$$

**End.**

**Output:** $F(\mathbf{x}) = \sum_{j=1}^{t} w_j h_j(\mathbf{x})$, where $\mathbf{w}$ is the solution of the primal problem (12) corresponding to the last dual solution.

In fact, $\mathbf{w}$ can also be solved as the dual variable of (19) by the interior-point methods.

For fast convergence in implementation, we employ a strong oracle as follows: $h_t = \arg\max_h \sum_{i=1}^{M} u_i y_i h(\mathbf{x}_i)$. However, the strong oracle is not a necessary condition. Actually any newly-added classifier just needs to violate the constraint that $\sum_{i=1}^{M} u_i y_i h_t(\mathbf{x}_i) \leq r$ to ensure the decrease of the objective value of the dual problem (18).

## 5 Experiments

In this section, we empirically demonstrate the performance of MCBoost and make a comparison with typical algorithms: AdaBoost, L2Boost [3], LPBoost [13], AdaBoost-CG [17] and MDBoost [18]. As was stated in [11] the complexity of base classifiers affects generalization performance, thus we use decision stumps as base classifiers for ease of their complexity. In addition, we limit the value of $E$ to the interval (0, 1) and use the CV to determine its optimal value. When more than one values of $E$ reach an identical minimum validation error, we choose the largest one. As discussed in Section 3, MCBoost is actually equivalent to solve the following problem

$$\min_{\mathbf{w}} \sum_{i=1}^{M} (1 - y_i H_{i:}\mathbf{w})^2, \, s.t. \, \mathbf{w} \geq 0 \text{ and } \|\mathbf{w}\|_1 = \frac{1}{E}. \tag{20}$$

The constraint $\|\mathbf{w}\|_1 = 1/E$ can actually be equivalent to $\|\mathbf{w}\|_1 \leq 1/E$ because if the optimal solution satisfies $\|\mathbf{w}\|_1 \prec 1/E$, one can enlarge the weights of any pair $(h_l, h_{l'})$ with $h_l(x) = -h_{l'}(x), \forall x$ such that $\|\mathbf{w}\|_1 = 1/E$. Thus (20) is also equivalent to

$$\min_{\mathbf{w}} \sum_{i=1}^{M} (1 - y_i H_{i:}\mathbf{w})^2 + \lambda \|\mathbf{w}\|_1, \, s.t. \, \mathbf{w} \geq 0 \tag{21}$$

Naturally, (21) is changed into a $l_1$-norm regularized or lasso-penalized [20] version of L2Boost. It is known that such a regularization can lead to the sparsity of **w**, thus forming a sparsely-combined strong classifier by the selection for the base classifiers. It is worth pointing out that since multiple values of $E$ likely correspond to the same minimum validation error, we here choose its largest one for yielding a final sparse strong classifier.

In what follows, firstly let us illustrate MCBoost's classification behavior on a toy dataset with 800 points (Fig. 1 (top-left)), 60% of which is used for training and the remaining for test. The termination threshold $\varepsilon$ is set to $10^{-5}$ and the maximum number of iterations $N_{max}$ to 1000. The cumulative margin distributions for different values of $E$ are plotted in Fig. 1(top-middle). We can observe that an over large value ($>= 0.5$) of $E$ can result in a MD with large average margin and large margin variance. In this case, both the training and the test errors are relative high (Fig. 1(top-right)). Naturally, large average margin is indeed beneficial to generalization but should not be at the expense of the margins of over many training samples. Schapire's margin theory [9] has stated that a large MD is important to generalization performance but generally, a large MD does not necessarily guarantee a large average margin, what is important is that the MD can make the margins of most training examples large. As in Fig. 1(top-middle), an over large $E$ value makes the margins of over many training examples small. In contrast, an over small $E$ value results in a MD with both small average margin and small margin variance, and in this case, the training error can be small, even zero, but on the contrary, the test error is not ideal enough. Generally speaking, the margin of a training example can be regarded as a confidence measure of classification. Therefore, though small $E$ can produce small training error, its corresponding classification result tends to be not confident enough for which an overfitting can happen. As shown in Fig. 1(bottom-left), the number of iterations (or of base classifiers) decreases with $E$ increasing, which is consistent with the discussion above. In summary, we can conclude that an over large $E$ can yield a large average margin and small number of base classifiers but a large margin variance, a large training error and a large test error (i.e., underfitting). Contrarily, an over small $E$ can produce a small margin variance and a small training error but large number of base classifiers, a small average margin and a large test error (i.e., overfitting). In this experiment, an optimal MD is achieved when $E=0.3$.

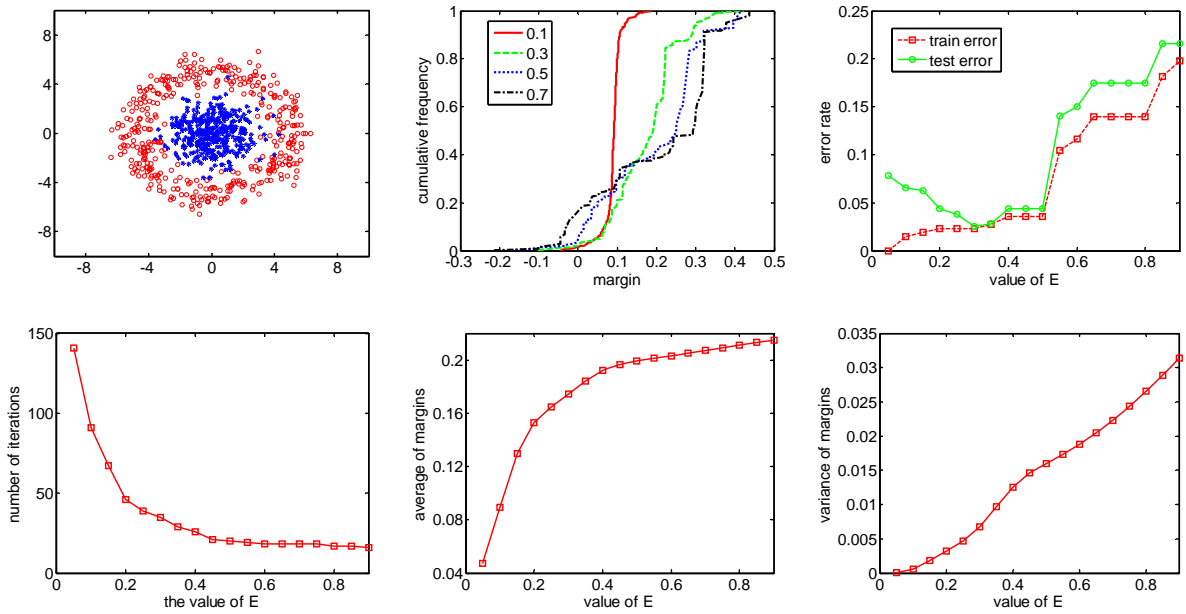

Fig.1. Toy data: (top-left) the input data; (top-middle) cumulative MDs for various values of $E$; (top-right) the training error and test error; (bottom-left) # of iterations (or of base classifiers combined); (bottom-middle) the average of margins; (bottom-right) the variance of margins.

The second experiment is specially designed to show AdaBoost's behavior in the viewpoint of MD. Here we select UCI's **breast-cancer** dataset on which AdaBoost yields an overfitting [18]. The data is randomly divided into two parts: 60% for training and 40% for test. As observed in Fig. 2, AdaBoost's MD changes with the iteration numbers and its average margin and margin variance decrease with the increase of the iteration number. Such a behavior is quite similar to that of MCBoost just as discussed before. As a result, AdaBoost can also be regarded as a MD controlled boosting and the number of its iterations exactly acts as a parameter to control the distribution.

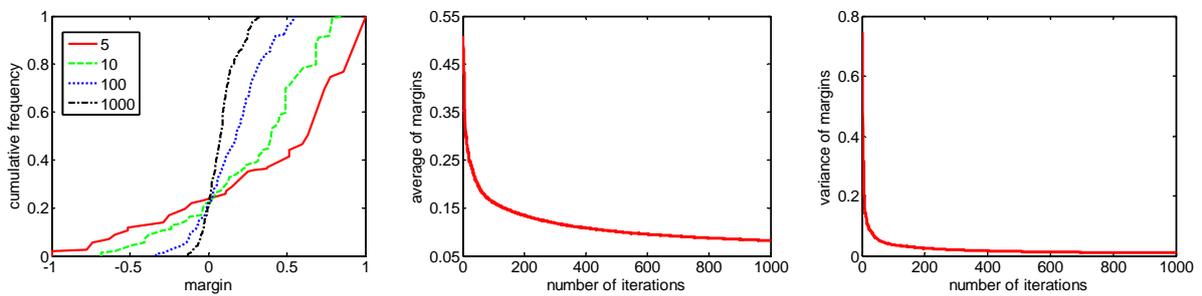

Fig. 2. MDs (left) of AdaBoost on breast-cancer with different iteration numbers. And the average (middle) and the variance (left) of the margins.

In the third experiment, we evaluate the empirical performance of our MCBoost on 13 UCI benchmark datasets [21] via comparing to the other five boosting algorithms: AdaBoost, L2Boost [3], soft-margin LPBoost [13], AdaBoost-CG [17] and MDBoost [18]. The comparing reason just with them lies in that 1) AdaBoost is the most popular algorithm in machine learning community; 2) L2Boost shares a similar motivation to our algorithm; 3) LPBoost is also a MD-controlled algorithm by maximizing the average of the k smallest margins [14]; 4) AdaBoost-CG and our MCBoost modify respectively the standard AdaBoost and L2Boost by fixing the sum of weak classifiers' coefficients and 5) MDBoost is the most recent MD-controlled algorithm which maximizes the average margin while minimizes the margin variance.

We use the CV to estimate the optimal numbers of boosting iterations in AdaBoost and L2Boost. Here the maximum iteration number is set to 1000 and we choose the smallest one that causes the optimal validation error. We adopt the same convergence threshold $\varepsilon$ of $10^{-5}$ for MCBoost, LPBoost, AdaBoost-CG and MDBoost and the same experiment setting as that of [18], i.e., each dataset is randomly split into three subsets where 60% for training, 20% for validation and the remaining 20% for test. For those large (ringnorm, twonorm and waveform) datasets, 10% is used for training, 30% for validation, and the rest 60% for test. We repeat the experiments for 10 times for the three large datasets and for 50 times for the other 10 datasets. All the results are reported in Table 1.

Table 1: Test errors of MCBoost (MC), AdaBoost (AB), L2Boost (L2B), AdaBoost-CG (AB-CG) and MDBoost (MD) (best method in bold face)

| Dataset  | MC           | AB           | L2B          | LP        | AB-CG        | MD           |
|----------|--------------|--------------|--------------|-----------|--------------|--------------|
| banana   | **26.5±1.2** | 27.1±1.5     | 26.7±1.3     | 32.2±2.1  | 28.0±1.0     | 27.7±0.7     |
| b-cancer | **27.4±5.1** | 28.5±5.1     | 28.4±5.2     | 34.0±7.2  | 29.4±5.7     | 28.5±4.4     |
| diabetes | **23.3±3.2** | 23.6±3.6     | 24.9±3.8     | 26.4±3.7  | 24.5±3.7     | 23.7±3.9     |
| f-solar  | **33.1±3.0** | **33.1±3.0** | **33.1±3.0** | 34.1±3.7  | 34.0±3.4     | 34.0±3.5     |
| german   | **24.4±3.1** | 24.9±2.8     | 25.1±2.6     | 30.5±3.5  | 25.5±3.0     | 25.6±2.8     |
| heart    | 16.7±3.9     | 19.3±4.4     | 18.1±5.0     | 18.5±5.7  | 17.1±4.7     | **16.1±4.2** |
| image    | 3.4±0.7      | 4.5±1.1      | 3.8±0.8      | 3.2±0.9   | **3.1±0.9**  | 3.3±1.0      |
| ringnorm | **5.1±0.4**  | 5.7±0.3      | 6.9±0.3      | 5.4±0.3   | 5.3±0.3      | **5.1±0.4**  |
| splice   | **7.4±0.8**  | 7.8±0.8      | 8.3±0.7      | 10.2±2.3  | 8.9±1.3      | 8.2±1.0      |
| thyroid  | **7.2±3.1**  | 8.2±3.8      | 8.4±3.9      | 7.8±5.1   | 7.8±4.5      | 7.6±4.9      |
| titanic  | 22.5±1.4     | 22.4±1.1     | **22.3±1.2** | 22.5±1.0  | 22.4±0.6     | 22.5±0.6     |
| twonorm  | **3.5±0.3**  | 4.2±0.3      | 5.1±0.3      | 4.3±0.3   | 4.1±0.4      | **3.5±0.2**  |
| waveform | 12.5±0.8     | **12.3±0.6** | 13.9±0.8     | 12.7±0.6  | 12.4±0.9     | 12.8±0.9     |

Table2: Comparison of MCBoost with AdaBoost, L2Boost, LPBoost, AdaBoost-CG and MDBoost using Wilcoxon signed-rank test (one-tail).

|  | AB | L2B | LP | AB-CG | MD |
|---|---|---|---|---|---|
| $z$-statistic | 2.8 | 2.92 | 2.92 | 2.67 | 2.17 |

Our MCBoost shows outstanding performance compared with the other five algorithms. As can be seen, MCBoost wins on most datasets (9 of 13) in terms of generalization error. The Wilcoxon signed-rank test [22] is used to show the superiority of MCBoost quantitively. And the $z$-statistics results shown in Table 2 are always in favor of our MCBoost. If we set the confidence level to 95% which corresponds to a critical value of 1.645, we have the evidence to believe that MCBoost outperforms all the other compared algorithms.

Next let us compare the computational complexity between MCBoost and MDBoost [18], they both need to solve a QP in each iteration but their difference in solving is that the quadratic term matrix in MCBoost is an identity while in MDBoost, a pseudo-inverse of a positive semi-definite matrix, hence the complexity of MCBoost is lower than that of MDBoost.

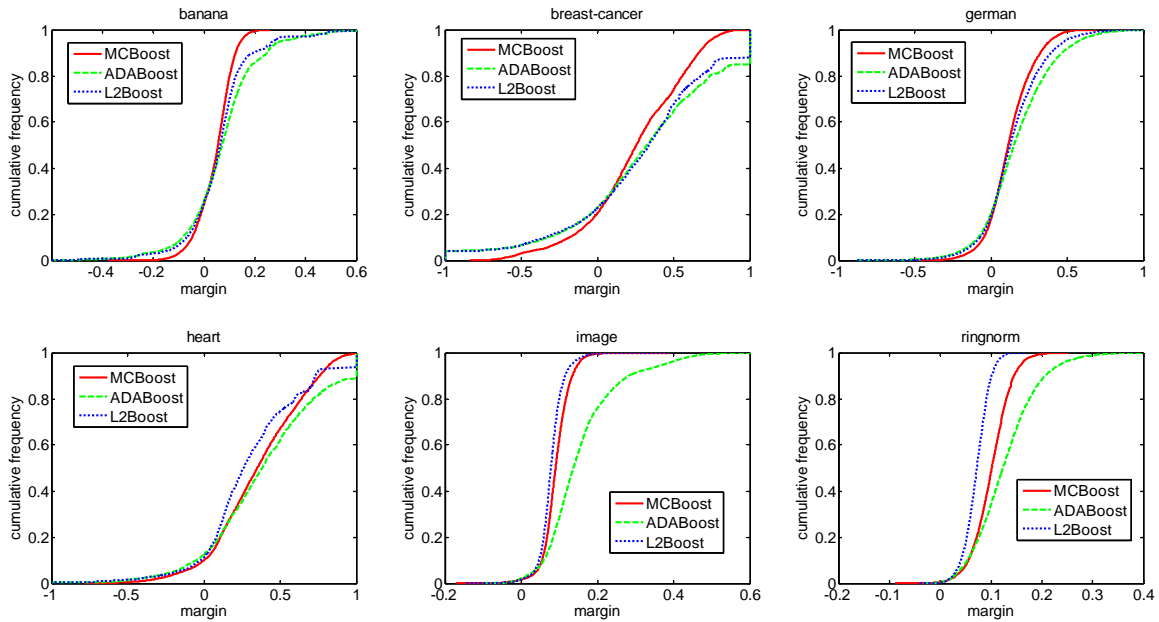

Fig. 3. Cumulative MDs for MCBoost, AdaBoost and L2Boost on six different datasets.

Since our algorithm is based on the MD of training examples, we also exhibit the cumulative MDs on the six datasets in Fig.3. Here we only present the results of MCBoost, AdaBoost and L2Boost just for illustration. As can be seen that in most cases, MCBoost's MDs can be characterized by a smaller variance. And typically MCBoost has a shorter tail over incorrect predictors. As have been discussed in [18] and [19] that a small margin variance may be in favor of generalization performance and our experiments support this viewpoint again.

Finally, we examine the convergence behavior of our MCBoost (Fig. 4), it can be observed that MCBoost converges much faster than AdaBoost, which is because our MCBoost is a totally corrective boosting algorithm, and the weights of its all past classifiers are updated in each iteration. While AdaBoost performs the coordinate gradient descent (CGD) in function space [2] and thus typically has a slow convergence. For AdaBoost, overfitting appears on both breast-cancer and German, thus its iteration number has to be tuned to avoid the overfitting.

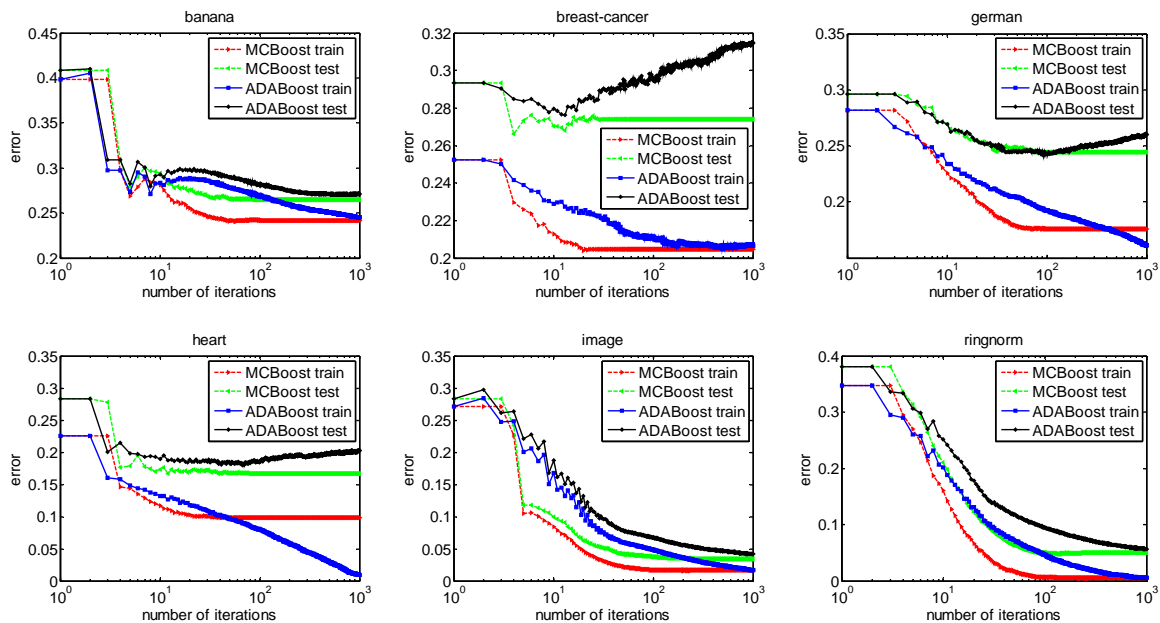

Fig.4. Training errors and test errors of MCBoost and AdaBoost on six datasets. The result of each plot is on an average of 50 times repeated experiments (10 times for ringnorm).

## 6 Conclusion

In this paper, we propose a novel MCBoost algorithm which *directly* controls the MD by minimizing the sum of the squared differences between the margins and an introduced adjustable (desired margin) parameter, and then use the CV to determine the optimal margin parameter. The algorithm is designed based on the CG which ensures fast convergence, and thus just small number of weak classifies is involved in a finally-formed strong classifier. We compare our MCBoost with the other five typical boosting algorithms and empirically illustrate the superiority of MCBoost in generalization ability.

CV is a simple but effective method to determine a good MD, while many validation values for $E$ are needed in order to get a smaller validation error. Therefore one of our future works is to find a search strategy to reduce the number of validation values of *E*.

**Acknowledgement**

This work is supported by National Natural Science Foundations of China under Grant Nos. 60973097 and 61035003.